# HybridNets: End-to-End Perception Network

**VT Dat[a] NVH Bao[a] PD Hung[a]**

[a] *FPT University, Hoa Lac High Tech Park, Hanoi, 10000, Vietnam*

datvthe140592@fpt.edu.vn, baonvhhe141782@fpt.edu.vn, hungpd2@fe.edu.vn

**Abstract**— End-to-end Network has become increasingly important in multi-tasking. One prominent example of this is the growing significance of a driving perception system in autonomous driving. This paper systematically studies an end-to-end perception network for multi-tasking and proposes several key optimizations to improve accuracy. First, the paper proposes efficient segmentation head and box/class prediction networks based on weighted bidirectional feature network. Second, the paper proposes automatically customized anchor for each level in the weighted bidirectional feature network. Third, the paper proposes an efficient training loss function and training strategy to balance and optimize network. Based on these optimizations, we have developed an end-to-end perception network to perform multi-tasking, including traffic object detection, drivable area segmentation and lane detection simultaneously, called HybridNets, which achieves better accuracy than prior art. In particular, HybridNets achieves **77.3 mean Average Precision** on Berkeley DeepDrive Dataset, outperforms lane detection with **31.6 mean Intersection Over Union** with **12.83 million** parameters and **15.6 billion** floating-point operations. In addition, it can perform visual perception tasks in real-time and thus is a practical and accurate solution to the multi-tasking problem. Code is available at https://github.com/datvuthanh/HybridNets.

## 1. INTRODUCTION

### 1.1. Background

Recent advances in embedded systems' computational power and neural networks' performance have made autonomous driving an active field in computer vision. Ideally, to create a vehicle capable of driving itself is to feed it with every bit of information available in its immediate surroundings. However, unlike conventional thinking, lidar and radar are not required to create an accurate perception field for intelligent vehicles. From time to time, it has been shown that such vehicles can make relatively good driving decisions with just the assistance of a single camera attached to the front. There is a general consensus that the three most critical tasks in guiding intelligent vehicles are: traffic object detection, drivable area segmentation, and lane line segmentation.

Each one of these tasks has got its state-of-the-art networks, including but not limited to SSD [14], YOLO [21] for object detection; UNet [23], SegNet [1], ERNet [10] for semantic segmentation; LaneNet [29] and SCNN [19] for lane line detection. Still, passing an image through three different networks creates unreasonable latency. Many researchers (MultiNet [28], DLT-Net [20], YOLOP [30]) have thought about combining the networks into a simple encoder-decoder architecture, where the backbone and neck generate context for three different heads to process. The architecture can be improved even further with proper selection of the feature extractor and fusing lane line with drivable area into one segmentation head. This experiment achieves the highest recall of 92.8% and segmentation IoU of 70.8%, outperforming existing multi-task networks on the challenging BDD100K dataset [31], as shown qualitatively in Figure 1.

Improvements are made upon the excellent multi-scale feature fusion BiFPN in EfficientDet [26], together with an EfficientNet [27] backbone pre-trained on ImageNet with its balanced trade-off between accuracy and computational overhead. A BiFPN decoder is constructed to utilize existing multi-scale features into the newly designed segmentation head. For an input resolution of 640x384, the entire network comes in at 15.6 BFLOPS on 12.83M parameters, comparable to the latest multi-task network YOLOP at 18.6 BFLOPS on 7.9M parameters. A multi-stage learning strategy is employed to help with the convergence of multiple loss functions [5].

To finetune even further, we also tinker with anchor box generation in this study [22]. Because anchor boxes theoretically cannot be generalized well for every dataset, nevertheless having a significant impact on the performance of one-stage detectors, we empirically choose the best possible aspect ratios and scales for the driving dataset BDD100K, where objects vary from large upfront trucks to tiny further cars.

To sum it up, the main contributions of this research are:

1. HybridNets, an end-to-end perception network, achieving outstanding results in real-time on the BDD100K dataset.
2. Automatically customized anchor for each level in the weighted bidirectional feature network, on any dataset.
3. An efficient training loss function and training strategy to balance and optimize multi-task networks.

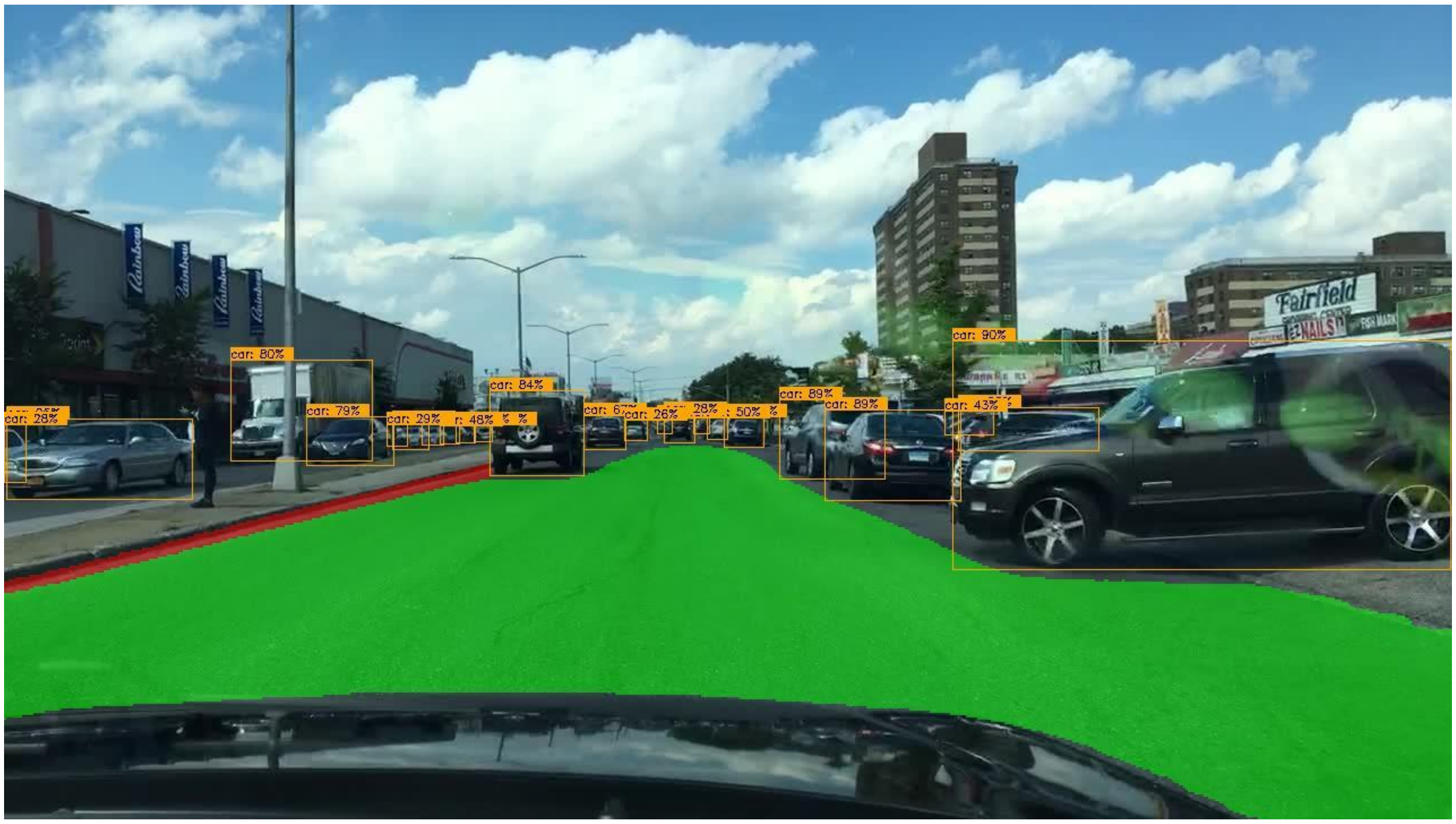

**Fig. 1.** Results for inference of HybridNets. Our proposed network performs three tasks, including traffic object detection, drivable area segmentation and lane line detection. The green areas indicate the drivable area, the blue lines are the lane lines, and the orange boxes are the traffic objects.

## 1.2. Related Works

This section will review some of the best networks in each respective task, then conclude with the latest multi-task networks to emphasize the strength of this unified architecture.

Current developments in improving detectors' performance have nearly split the area into two distinct branches: region-based and one-stage detectors. While region-based methods are more accurate, one-stage detectors gained more attraction due to their efficiency in embedded systems with limited hardware constraints. When FPN came about, it initially supported RPNs by providing a top-down pathway to construct higher resolution layers from a semantic-rich layer [11]. Then BiFPN officially showed the performance boost of bidirectional feature fusion to one-stage detectors. They can now take in multiple scales of the feature map in just one pass, alleviating the apparent weakness of YOLOs and the like.

Semantic segmentation has also made remarkable steps with deep-learning instead of the old-fashioned segmentation algorithms. FCN [25] sparked the flame with the first fully convolutional segmentation network. From then on, researchers have found various ways to improve the performance, such as encoder-decoder architecture with UNet [23], the pyramid pooling module of PSPNet [32], or even semi-supervised learning based on generative adversarial networks [6]. SSN [18] incorporated conditional random field units in the post-processing stage to increase segmentation performance. Many data augmentation techniques have been tested throughout to enhance the learning generalization of road detection networks [17]. Image analysis is still being explored in segmenting road scenes [9].

Traditional lane line detections algorithms have been in wide use until recently, a notable algorithm being Hough transform [33]. Then LaneNet [29] proposed individual lane lines as instances to be segmented. Spatial CNN [19] preferred slice-by-slice convolutions over deep layer-by-layer convolutions, emphasizing objects with heavy spatial relationships but barely noticeable appearances, such as poles, traffic lights, or lane lines. ENet-SAD [8] created self attention distillation, a technique allowing models to self-learn. It works by using attention maps generated in earlier training points as a form of supervision for later, surpassing SCNN by a large margin.

Many published papers attempted to combine perception tasks into a unified network. Mask R-CNN [7] inherited RPN from Faster R-CNN while adding a third output branch for object mask, enabling the parallelization of object detection and instance segmentation. BlitzNet [4] also showed that object detection and semantic segmentation could benefit from each other. LSNet [13] came with a novel loss function named cross-IoU to add pose estimation into the output. MultiNet put forward the encoder-decoder structure, allowing DLT-Net to design special shared tensors between decoder heads for mutual information streams. Not long after, YOLOP became the first real-time state-of-the-art on the BDD100K dataset on three perception tasks: vehicle detection, drivable area, and lane line segmentation. However, the two similar segmentation heads left room for the obvious optimization task of reducing them to a single better-performing one. As hardware constraint is also of utmost importance to the application of any real-time decision-making network, model scaling must be taken into consideration.

# 2. METHODOLOGY

## 2.1. Network Architecture

Based on these challenges, this research has proposed an end-to-end network architecture that can multi-task named HybridNets. As shown in Figure 2, our one-stage network includes one sharing encoder and two separated decoders to solve distinct tasks. The resolution of each feature map level $P_i$ represents a feature level with resolution of $1/2^i$ of the input images. For instance, if input resolution is 640x384, the $P_2$ represents feature level 2 ($640/2^2$, $384/2^2$) $= (160, 96)$, while $P_7$ represents feature level 7 with resolution $(5, 3)$.

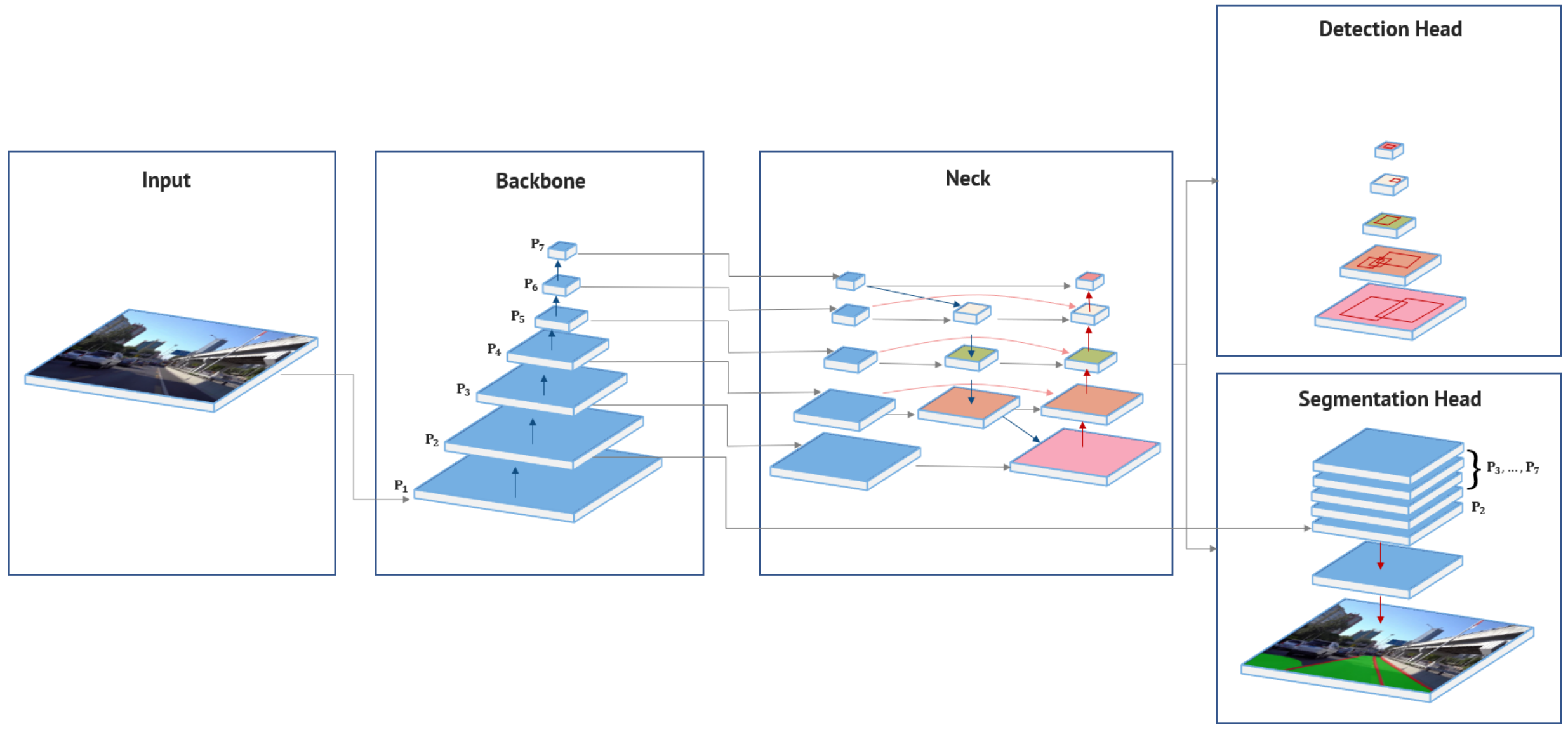

**Fig. 2.** HybridNets Architecture has one encoder: backbone network and neck network; two decoders: Detection Head and Segmentation Head. The backbone network generated 5

feature maps from $P_1$ to $P_5$. By down-sampling the feature map $P_5$,

we obtain two feature maps $P_6$ and $P_7$.

### 2.2. Encoder

The feature extracting, serving as a backbone, is an essential part of the model that can help a variety of networks achieve excellent performance to in various tasks. Many modern network architectures currently reuse networks that have good accuracy in the ImageNet dataset to extract features. Recently, EfficientNet showed high accuracy and efficient performance over existing CNNs, reducing FLOPs by orders of magnitude. We choose EfficientNet-B3 as the backbone, which solves the problem of network optimization by finding depth, width, and resolution parameters based on neural architecture search to design a stable network. Therefore, our backbone can reduce the computational cost of the network and obtain several vital features.

The feature maps from the backbone network are fed to the neck network pipeline. Multi-scale feature representation is the main challenge; FPN recently proposed a feature extractor design to generate multi-scale feature maps to obtain better information. However, the limitation of FPN is that information feature is inherited by a one-way flow. Therefore, our neck network uses a BiFPN module based on EfficientDet. BiFPN fuses feature at a different resolution based on cross-scale connection for each node by each bidirectional (top-down and bottom-up) path and adds weight for each feature to learn the importance of each level. We adopt the method to fuse features in our work.

### 2.3. Decoder

Each grid of the multi-scale fusion feature maps from the Neck network will be assigned nine prior anchors with different aspect ratios. Similar to YOLOv4 [2], this paper uses kmeans clustering [16] to determine anchor boxes. In addition, we chose 9 clusters and 3 different scales for each grid cell. In order to various feature map levels, this paper use scale constant to create bounding box priors that covers all regions from small to large. Thus, this proposed network can work well on complex dataset. The detection head will predict the offset of bounding boxes and the probability of each class as well as the confidence of the prediction boxes. This is described as:

$$b_x = \sigma(r_x) + c_x$$
$$b_y = \sigma(r_y) + c_y$$
$$b_w = c_w e^{r_w}$$
$$b_h = c_h e^{r_h}$$

Where $r_x, r_y, r_w, r_h$ is the center, width and height of each bounding box, respectively from network prediction. Each anchor box has a center $c_x, c_y$, width $c_w$ and height $c_h$.

Segmentation head has 3 classes for output, which are background, drivable area and lane line. This paper keeps 5 feature levels $\{P_3, ..., P_7\}$ from Neck network to segmentation branch. First, this paper up-samples each level to have the same output feature map with size $(\frac{W}{4}, \frac{H}{4}, 64)$. Second, feeding $P_2$ level to convolution layer to have the same feature map channels with other levels. Then, we combine them to obtain a better feature fusion by summing all levels. Finally, we restore the output feature to the size $(W, H, 3)$, representing the probability of each belonging pixel class. This research scales feature maps to the size of $P_2$ level, because $P_2$ level is a strongly semantic feature map. Additionally, we feed $P_2$ feature map from backbone network which represents low-level feature into the final feature fusion that helps network improve output precision, as shown in Figure 3.

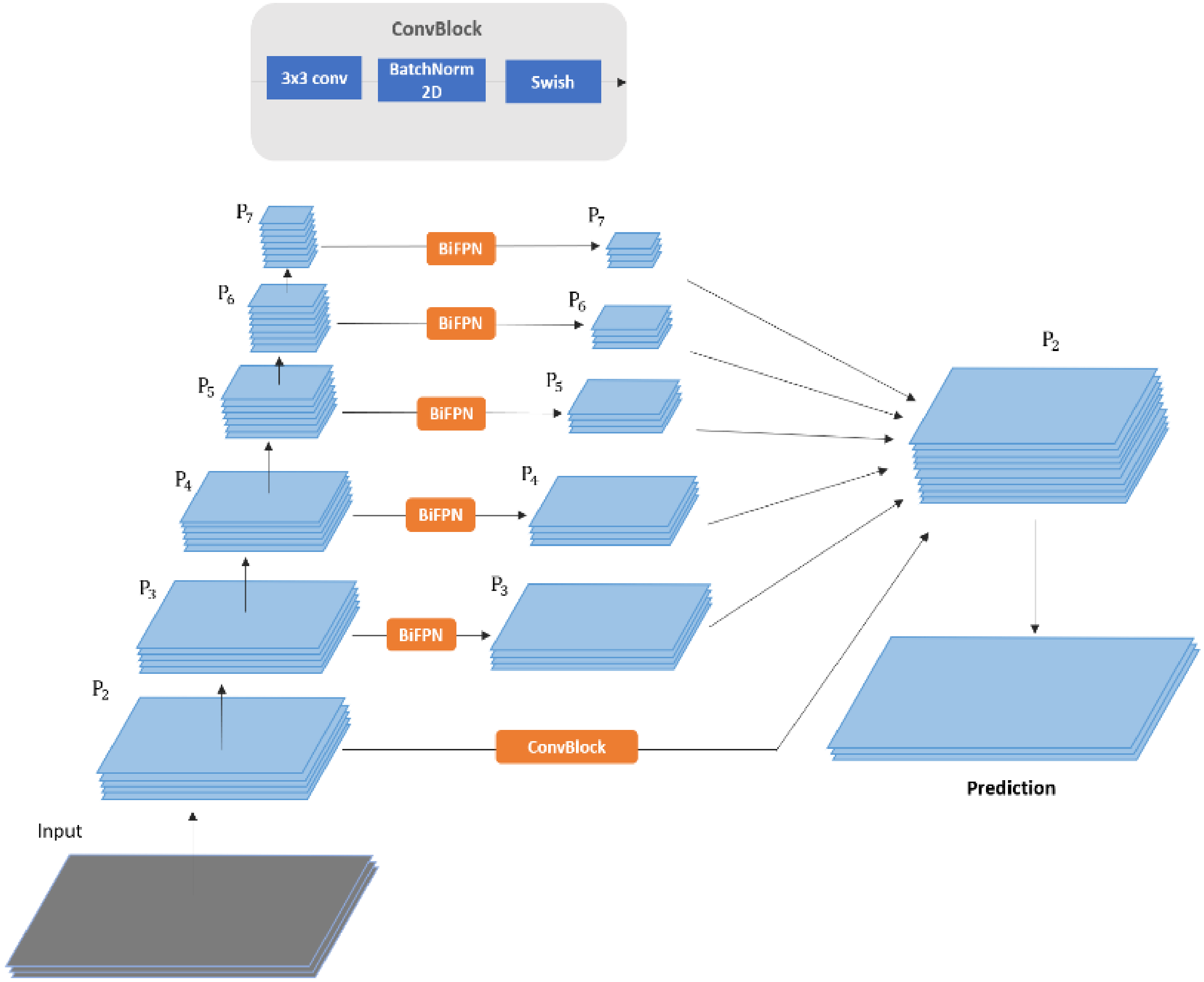

**Fig. 3.** The Segmentation branch of HybridNets architecture.

## 2.4. Loss Function and Training

This paper used multi-task loss to train end-to-end network. Equation 2 expressed the total loss function by summing of two parts.

$$L_{all} = \alpha L_{det} + \beta L_{seg}$$

Where $\alpha$ , $\beta$ are tuning parameters to balance the total loss, $L_{det}$ is the loss for object detection task and $L_{seg}$ is the loss for segmentation task, the formulation can be written as follow

$$L_{det} = \alpha_1 L_{class} + \alpha_2 L_{obj} + \alpha_3 L_{box}$$

$\mathcal{L}_{class}$ and $\mathcal{L}_{obj}$ are focal loss [12], which is implemented for classifying class and the confidence of objects, respectively. The focal loss reduces the slope of loss function and focuses on misclassified examples. $\mathcal{L}_{box}$ is computed by smooth L1 loss, which takes absolutely between the predicted box and ground truth box, can be expressed as

$$\text{smooth}_{L1}(\text{x}) = \begin{cases} \delta_1 x^2 \quad if \quad x < \delta_2 \\ x - \delta_1 \end{cases}$$
$$\text{x} = \delta b_p \cdot \left( \left| b_x - \hat{b}_x \right| + \left| b_y - \hat{b}_y \right| + \left| b_w - \hat{b}_w \right| + \left| b_h - \hat{b}_h \right| \right)$$

Where $\hat{b}$ is the prediction of bounding box and $b$ is the ground truth, and $b_p$ is determined a positive label has been assigned to a grid cell. In this paper, we force size some anchor boxes to the regression network can learn smoothly, $b_p$ can be written as

$$b_p = \begin{cases} 1, \quad if \quad IoU(c_i, b_j) \ \geq \ 0.5 \quad i = 1, ..., \sum_{k=1}^{5} n_k m_k; j = 1, ..., p \\ 0, \quad otherwise \end{cases}$$

Where $c_i$ is the anchor box $i^{th}$, the total of anchor boxes is combining of each feature map level with $n_k, m_k$ is the resolution of feature map, and $p$ is the total of ground truth bounding boxes of each input image. Next $\mathcal{L}_{seg}$ is multiclass hybrid loss that is utilized for multi-class segmentation of background, drivable area and lane line. Small object segmentation is a challenge in semantic segmentation caused by imbalanced data distribution. Therefore, this paper combines $\mathcal{L}_{Tversky}$ Tversky loss [24] and $\mathcal{L}_{Focal}$ Focal loss [12] to predict the class to which a pixel belongs. $\mathcal{L}_{Tversky}$ performs well at class-imbalanced problems and optimizes the maximization of score, whereas $\mathcal{L}_{Focal}$ aims to minimize the classification error between pixels and focuses on hard labels.

$$TP_p(c) = \sum_{n=1}^{N} p_n(c) g_n(c)$$

$$FN_p(c) = \sum_{n=1}^{N} (1 - p_n(c)) g_n(c)$$

$$FP_p(c) = \sum_{n=1}^{N} p_n(c)(1 - g_n(c))$$

$$\mathcal{L}_{seg} = \mathcal{L}_{Tversky} + \lambda \mathcal{L}_{Focal}$$

$$\mathcal{L}_{Tversky} = C - \sum_{c=0}^{C-1} \frac{TP_p(c)}{TP_p(c) + \varphi FN_p(c) + (1-\varphi) FP_p(c)}$$

$$\mathcal{L}_{Focal} = -\lambda \frac{1}{N} \sum_{c=0}^{C-1} \sum_{n=1}^{N} g_n(c)(1 - p_n(c))^{\gamma} \log(p_n(c))$$

Where $TP_p(c)$ , $FN_p(c)$ , $FP_p(c)$ are true positives, false negatives and false positives for class $N$ , $p_n(c)$ is the predicted probability for pixel $n$ belonging to class $c$ , $g_n(c)$ is the ground truth for pixel $n$ being in class $c$ . $C$ is the number of classes and $N$ is the total number of pixels in the input image.

During training, this paper did several experiments to finetune a lot of hyperparameters and suitable architecture networks. Training from an end-to-end approach will cost computation and training time. In addition, several optimization algorithms have also been experimented. Therefore, to compress the training time and optimize hyperparameters, we construct a training strategy in order to train the model step by step and quickly transform the experiments. Algorithm 1 illustrates the strategy of our training method.

**Algorithm 1**. HybridNets training stage. First, we only train Encoder and Detection Head as object detection task. Second, we freeze the Encoder, Detection head and unfreeze parameters from Segmentation Head. Finally, the final network is trained jointly for all tasks.

**Input:** Target end-to-end network $\mathsf{F}$ with parameter group

$\Theta = \{\theta_{enc}, \theta_{det}, \theta_{seg}\}$ ;

Training dataset $\mathsf{T}$ ;

Threshold for convergence $\gamma = \{\gamma_1, \gamma_2, \gamma_3\}$ ;

Total loss function $\mathsf{L}$ ;

Pivot strategy $\mathsf{P} = \{\{\theta_{enc}, \theta_{det}\}, \{\theta_{seg}\}, \{\theta_{enc}, \theta_{det}, \theta_{seg}\}\}$

**Output:** Proposed network: $\mathsf{F}(\mathsf{X}, \Theta)$

**procedure** Train $(\mathsf{F}, \mathsf{T})$

    **for** $i$ = 0 to length $(\mathsf{P})$ - 1

        $\Theta \leftarrow \Theta \cap \mathsf{P}[i]$ // Freeze parameters

        **repeat**

            Sample a mini-batch $(\mathbf{x}_m, \mathbf{y}_m)$ from training dataset $\mathsf{T}$ .

            $\ell \leftarrow \mathsf{L}_{all}(\mathsf{F}(\mathbf{x}_m; \Theta), \mathbf{y}_m)$

            $\Theta \leftarrow \arg\min_\theta \ell$

        **until** $\ell < \gamma[i]$

        **if** $i$ < length $(\mathsf{P})$ - 1 **then**

            $\Theta \leftarrow \Theta \cup \mathsf{P}[i+1]$

        **endif**

    **end for**

**end procedure**

Train $(\mathsf{F}, \mathsf{T})$

**return** Proposed network $\mathsf{F}(\mathsf{X}, \Theta)$

# 3. EXPERIMENTATION AND EVALUATION

## 3.1. Experiment Settings

The Berkeley DeepDrive Dataset (BDD100K) is used in training and validating the model. Since the test labels of 20K images are unavailable, we opt to evaluate on the validation set of 10K images. The dataset for three tasks is prepared according to existing multi-task networks trained on BDD100K to aid in comparison. Of all the ten classes in object detection, only {car, truck, bus, train} is selected and merged into a single class {vehicle} since DLT-Net and MultiNet can only detect vehicles. Two segmentation classes {direct, alternative} are also merged into {drivable}. We follow the practice of calculating two lane line annotations into a central one, dilating the annotations in training set to 8 pixels while keeping validation set intact [8]. Images are resized from 1280x720 to 640x384 due to three main reasons, in order of importance: respecting the original aspect ratio, maintaining a good trade-off between performance and accuracy, and making sure the dimensions are divisible by 128 for BiFPN. Basic augmentation techniques such as rotating, scaling, translating, horizontal flipping, and HSV shifting are used. Mosaic augmentation, first introduced in YOLOv4 with great results [2], is utilized while training detection head specifically.

We jump-start the model by using EfficientNet-B3 weights pre-trained on ImageNet. The custom anchor box settings found automatically have scales of $(2^{0}, 2^{0.7}, 2^{1.32})$ and ratios of $[(0.62, 1.58), (1.0, 1.0), (1.58, 0.62)]$. The chosen optimizer is AdamW [15] with $\gamma = 1e^{-3}, \beta_1 = 0.9, \beta_2 = 0.999, \xi = 1e^{-8}, \lambda = 1e^{-2}$. When the model stucks around for 3 epochs, learning rate is decreased tenfold. For object detection, the model uses smooth L1 loss with $\delta_1 = 4.5, \delta_2 = 1/9$ for regression and focal loss with $\alpha = 0.25, \gamma = 2.0$ for classification. When matching anchor boxes to annotations, the model uses an IoU threshold of 0.5 for annotations larger than 100 pixels in area but only 0.25 for those smaller. We emphasize regression 4 times more than classification because one-class classification is easy to converge. For drivable area and lane segmentation, the model uses a combination of Tversky loss with $\alpha = 0.7, \beta = 0.3$ and Focal loss with $\alpha = 0.25, \gamma = 2.0$. We train with a batch-size of 16 on a RTX 3090 for 200 epochs.

### 3.2. Evaluation Metrics

On traffic object detection task, this proposed method uses mAP50. mAP50 is computed by the average of the Average Precision calculated for all the classes at single IoU threshold 0.5. Average Precision is the area under the precision-recall curve. This paper only evaluates one class, focusing on how good the proposed method can find all the positives. This paper set the lowest confidence and all bounding boxes is computed by mAP50. On semantic segmentation task, IoU metric is used to evaluate drivable area and lane line segmentation. To be more specific, this paper presents mIoU as average of IoU for each class and IoU metric for single class.

### 3.3. Cost Computation Performance

Table 1 compares HybridNets with other multi-task networks. Although our HybridNets has more extensive parameters (12.83M) than YOLOP (7.9M), the number of computations of HybridNets is lower than the compared networks. By adopting depth-wise separable convolutions [3], the computations are significantly reduced to **15.6 BFLOPs**. In addition, we have also compared the inference latency on V100 GPU FP16. Specifically, our V100 latency is the time processing of the model, not including preprocessing and NMS postprocessing. Compared to previous multi-networks, HybridNets are up to **1.4x** faster on GPU. Therefore, HybridNets can run in real-time on standard devices and embedded devices.

| Model | Params | FLOPs | Latency (ms) V100 |
|---|---|---|---|
| YOLOP | 7.9M | 18.6B | 52 |
| **HybridNets** | **12.83M** | **15.6B** | **37** |

**Table. 1.** Cost computation result for various multi-networks.
Params and FLOPs denote the number of parameters and the number of computations.
Latency is for inference with batch size 1.

### 3.4. Multi-task Performance

The second experiment presents results on three tasks, including traffic object detection, drivable area segmentation, and lane line segmentation. We present the vehicle detection results and compare them to six models on the BDD100K dataset.

| Model | Recall (%) | mAP50 (%) |
|---|---|---|
| MultiNet | 81.3 | 60.2 |
| DLT-Net | 89.4 | 68.4 |
| Faster R-CNN | 77.2 | 55.6 |
| YOLOv5s | 86.8 | 77.2 |
| YOLOP | 89.2 | 76.5 |
| **HybridNets** | **92.8** | **77.3** |

**Table. 2.** The comparison result on traffic object detection task. The experiment settings include confidence threshold of 0.001 and NMS threshold of 0.6. This paper mainly focuses on obtaining highest Recall IoU.

As listed in Table 2, HybridNets outperforms performance to previous networks on the BDD100K dataset. Our model achieves **3.6%** better recall and achieves the best mAP50 at **77.3%**. We can outperform all previous networks on recall and mAP50 metrics because our HybridNets can detect incredibly small objects ranging from 3 pixels to 10 pixels with input size (640,384,3) thanks to our automatically customized anchor aspect ratio and scale. Figure 4 illustrates the visualization of traffic object detection.

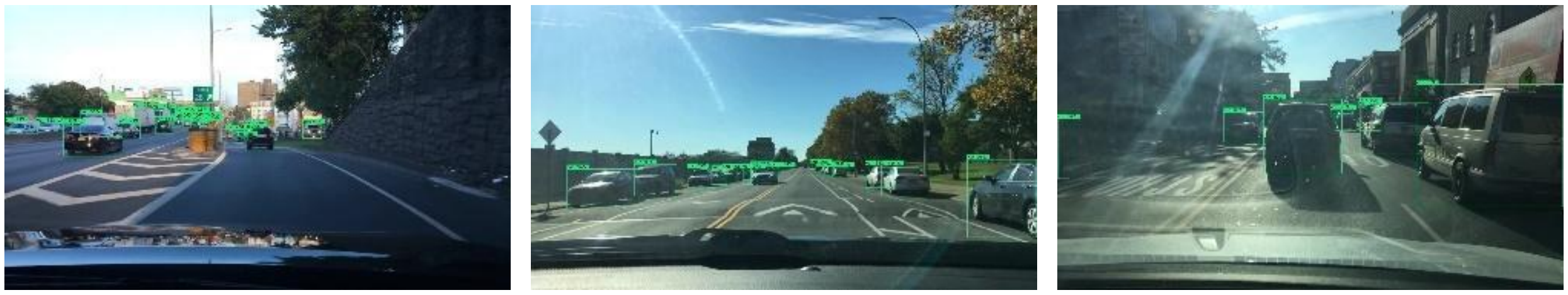

(a) Day-time result

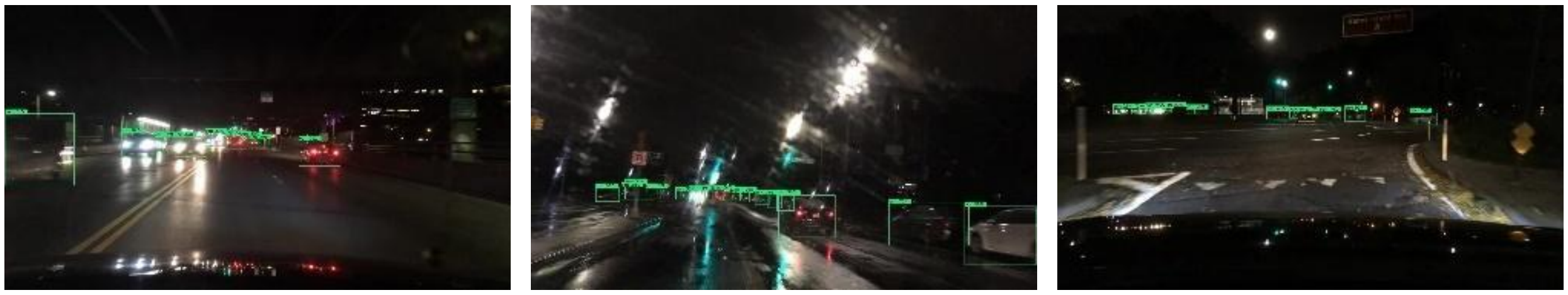

(b) Night-time result

**Fig. 4.** Visualization of the traffic object detection results of HybridNets. Fig. 4. (a) shows results in day-time series with different weather conditions such as clear, heat stroke and heat-wave. Fig. 4. (b) shows results in night-time series with different weathers such as cool and flurries.

As shown in Figure 5, our proposed architecture makes further improvement compared to YOLOP. Specifically, HybridNets detects small objects and large objects in traffic object detection task, whereas YOLOP has high False Negative score and detects wrong objects. In addition, HybridNets works well in various complex weather conditions and the bounding boxes are more accurate.

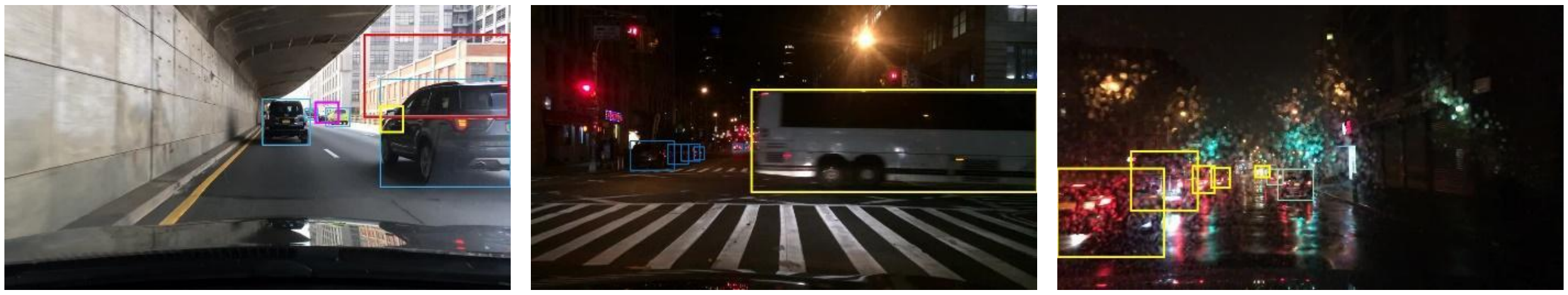

(a) YOLOP

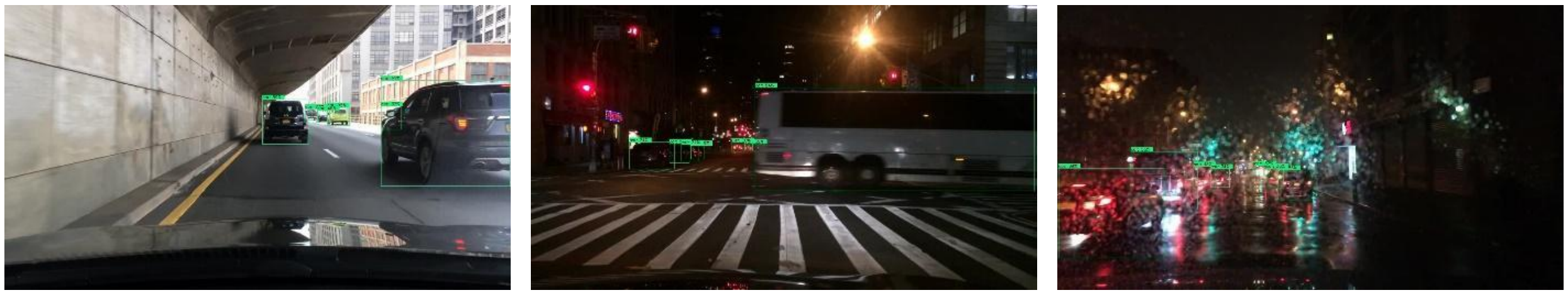

(b) HybridNets

**Fig. 5.** Comparison between YOLOP and HybridNets on Traffic Object Detection task. The first row shows issues of YOLOP and the second row shows the result of HybridNets. The red bounding boxes are the false positive, the yellow bounding boxes are the false negative and the purple bounding boxes are not accurate.

Next we evaluate the drivable area segmentation task. IoU metric is used to evaluate the segmentation performance of various networks.

| Model | Drivable mIoU (%) |
|---|---|
| MultiNet | 71.6 |
| DLT-Net | 71.3 |
| PSPNet | 89.6 |
| YOLOP | 91.5 |
| **HybridNets** | **90.5** |

**Table 3.** Performance comparison on drivable area segmentation task.

Table 3 shows the Drivable IoU of five networks. Our HybridNets achieves 90.5 % mIoU, pale in comparison to YOLOP (91.5%). We built a decoder network for multi-classes, whereas YOLOP constructed two decoders for specific tasks. Therefore, our HybridNets is more flexible and optimistic than theirs. Figure 6 visualizes the semantic segmentation output of drivable areas in various conditions. As shown in Figure 7, the comparison between HybridNets and YOLOP on Drivable Segmentation task, HybridNets is more accurate than YOLOP. To be more specific, YOLOP focuses on evaluating the pixel to which class it belongs, while needing to consider the intersection of bounding boxes. Therefore, the YOLOP model does not work well in harsh conditions such as night or areas with a lot of noise. Based on our Neck Network using BiFPN architecture, the information of different receptive fields is combined from various feature map levels with weighted parameters. Thus, HybridNets can improve the performance of drivable area segmentation task.

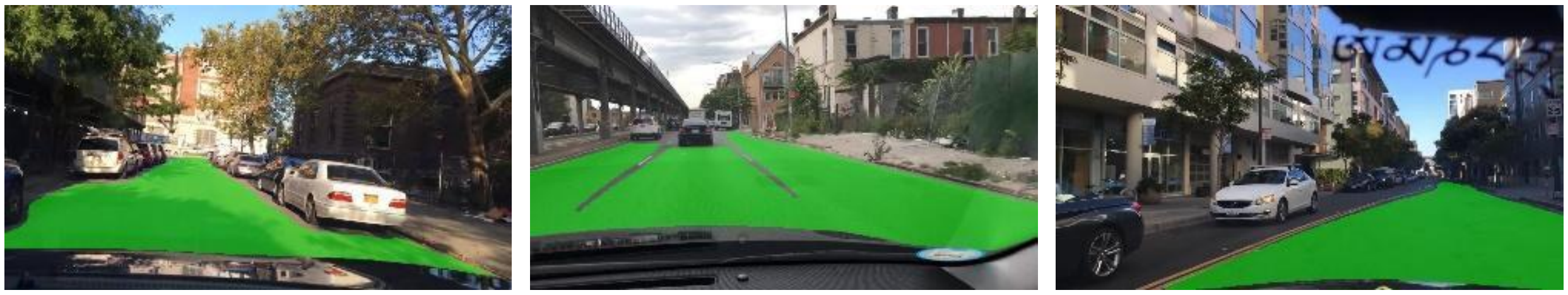

(a) Day-time result

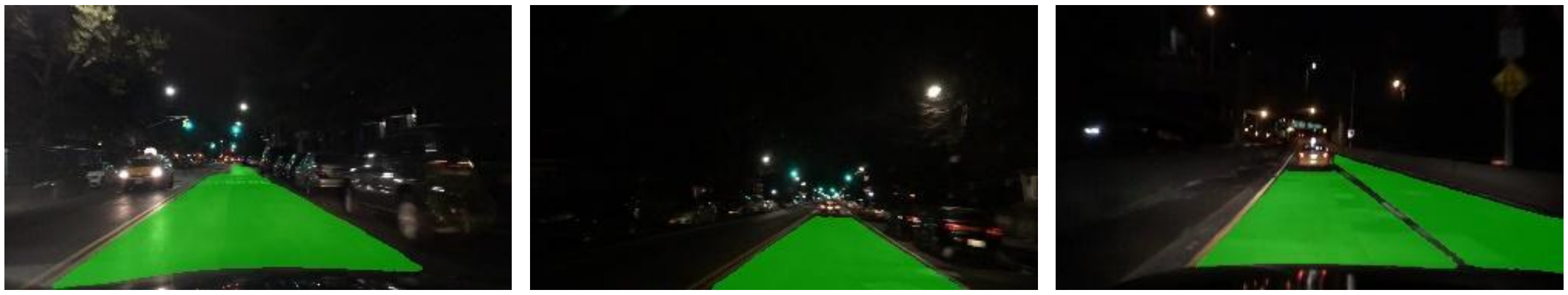

(b) Night-time result

**Fig. 6.** Visualization of the drivable area segmentation results of HybridNets. Fig. 6. (a) shows semantic segmentation results in day-time with various views. Fig. 6. (b) shows results in night-time series with various brightness views.

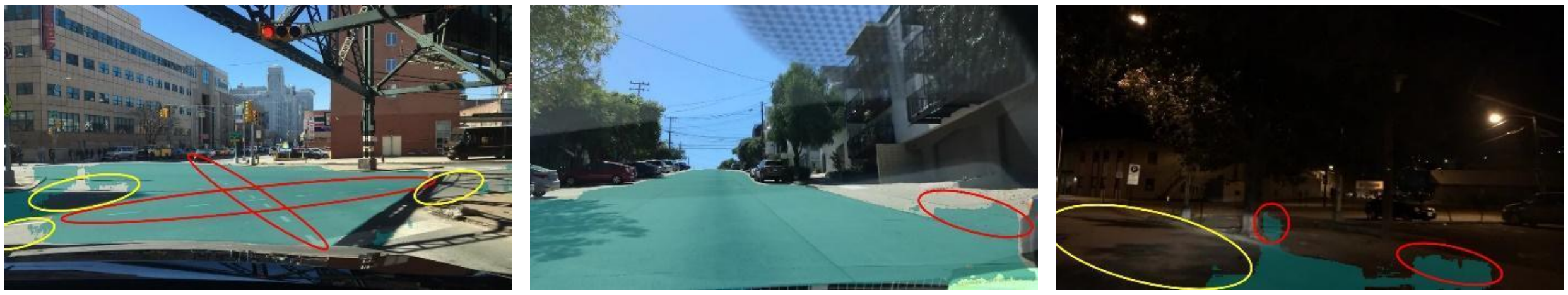

(a) YOLOP

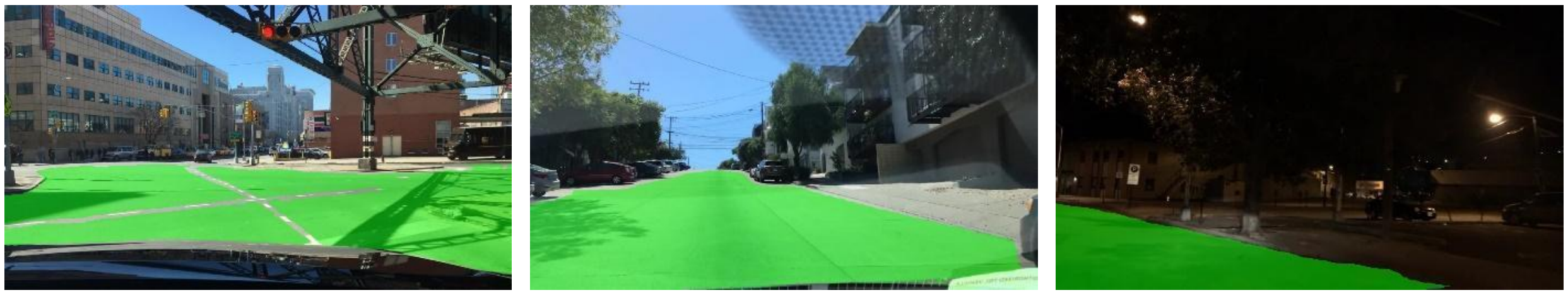

(b) HybridNets

**Fig. 7.** Comparison between YOLOP and HybridNets on Drivable Area Segmentation task. The first row shows the issue of mismatched pixels of YOLOP and the second row shows the result of HybridNets. The red regions are false positive and the yellow regions are false negative.

Finally, lane detection is one of the main challenges in autonomous driving. The evaluation metrics we use for lane detection are accuracy and IoU. As shown in Table 4, our HybridNets outperforms all previous models with accuracy **85.4** % and IoU **31.6** %. The proposed method works well in various complex weather conditions as shown in Figure 8. As shown in Figure 9, the lane detection results from YOLOP have mismatched pixels and less accuracy, whereas HybridNets works well on lane detection task. The lane lines from HybridNets are continuous and have high accuracy with less sparse supervisory. However, lane line has low IoU because of our approach in preprocessing the training dataset, making lane line annotation easier to learn with the drawback of suboptimal results. Thus, this paper has added another metric of accuracy to evaluate in a more objective and fair manner.

| Model | Accuracy (%) | Lane Line IoU (%) |
|---|---|---|
| Enet | 34.12 | 14.64 |
| SCNN | 35.79 | 15.84 |
| Enet-SAD | 36.56 | 16.02 |
| YOLOP | 70.50 | 26.2 |
| **HybridNets** | **85.4** | **31.6** |

**Table 4.** Performance comparison on lane detection task.

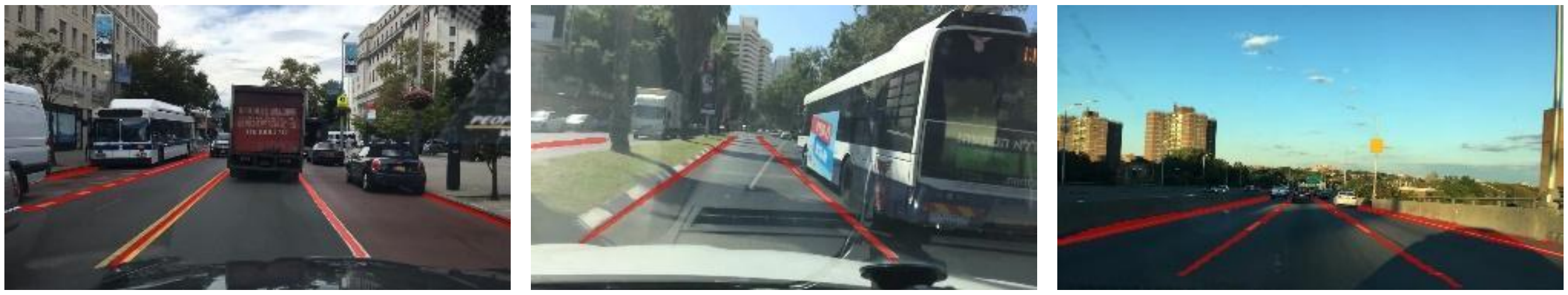

(a) Day-time result

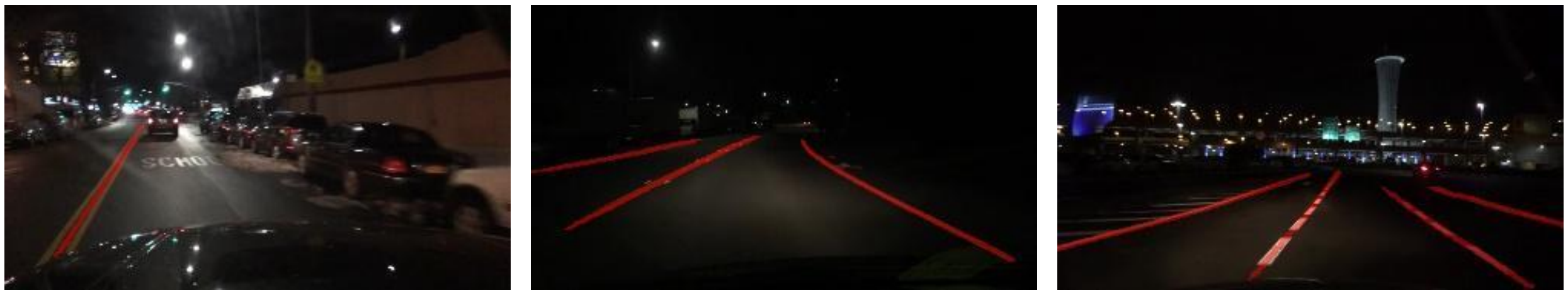

(b) Night-time result

**Fig. 8.** Visualization of the lane detection results of HybridNets. Fig. 8. (a) shows results in day-time series with various weather conditions. Fig. 8. (b) shows results in night-time series with various brightness views.

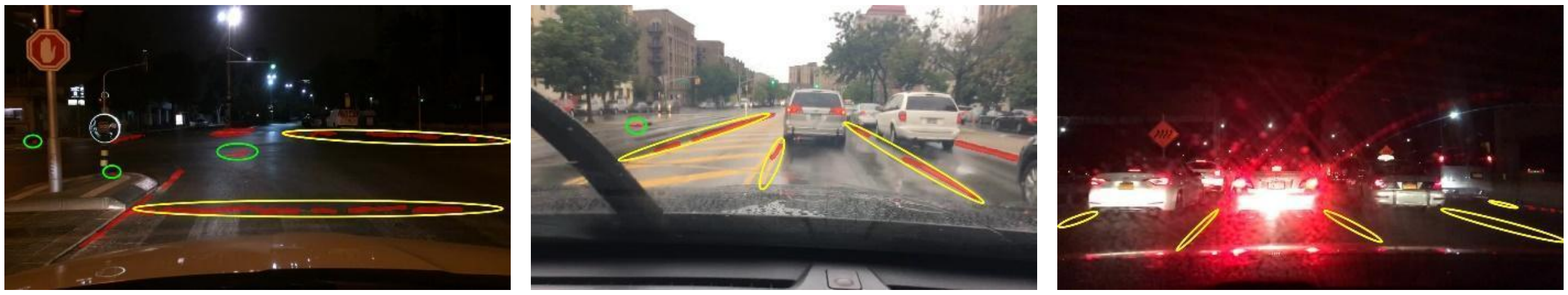

(a) YOLOP

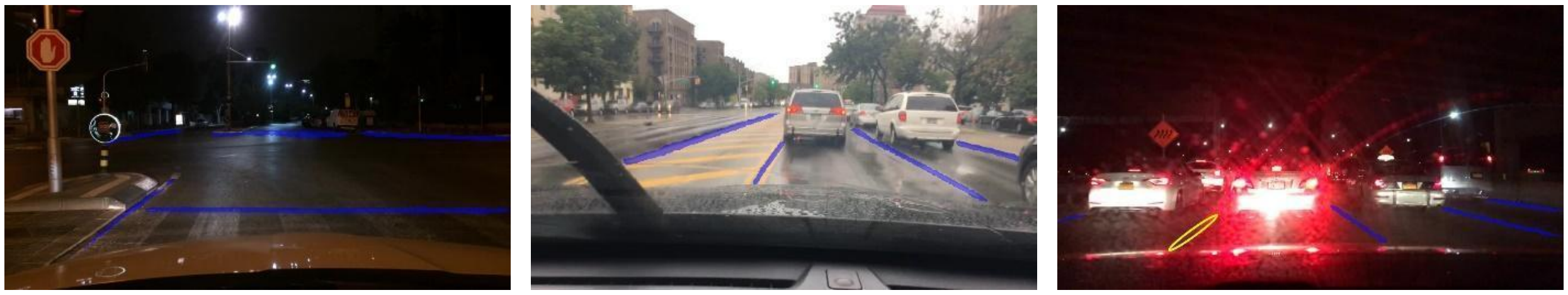

(b) HybridNets

**Fig. 9.** Comparison between YOLOP and HybridNets on Lane Detection task. The first row shows the issue of mismatched of YOLOP and the second row shows the result of HybridNets. The green regions are false positive and the yellow regions are false negative.

Figure 10 shows the results of HybridNets. The red lines are the lane lines, the green areas are the drviable area, and the orange bounding boxes are traffic objects. Our HybridNets has great performance in most scenarios. Based on the context structures, the drivable area provides information for the model to help train the model to converge faster. Moreover, each task provides context structure for other tasks. Therefore, our HybridNets can detect vehicles object easily, which challenges many other prior detection models. Therefore, the model can more easily predict traffic objects, which challenges many prior models. In general, our HybridNets works well in most complex scenarios such as severe reflective scenes and extreme weather conditions. However, our model is unable to adapt to the crossroads, the lane lines detection is broken and the drivable area is misjudged to be on the other side of the road in some cases.

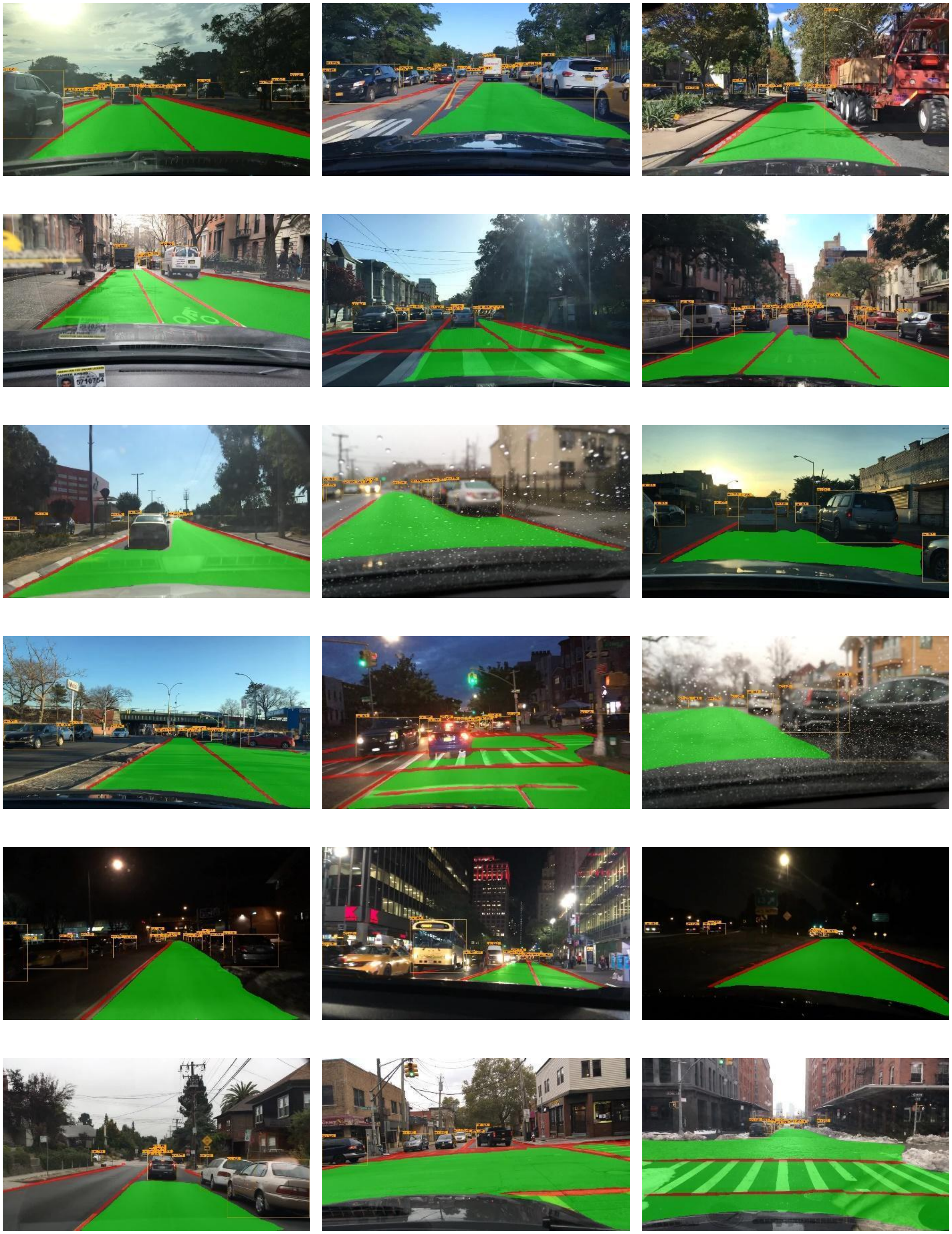

**Fig. 10.** Multi-task results using HybridNets. The red lines are the lane lines, the green areas are the drivable area, and the orange bounding boxes are traffic objects.

## 4. CONCLUSION AND PERSPECTIVE

In this paper, we systematically study network architecture design choices for multi-tasking, propose an efficient end-to-end perception network, customize automatic aspect ratios for each level in the weighted bidirectional feature network, and build efficient training loss function and training strategy to improve accuracy and performance. Based on these optimizations, we develop a new end-to-end multi-network, named HybridNets, which achieves better accuracy and efficiency than prior art across a broad spectrum of resource constraints. Most importantly, our network HybridNets achieves state-of-the-art accuracy with fewer FLOPS than previous multi-network models.

In future works, we would like to propose a robust network, which can perform many tasks related to perception and improve parameters and FLOPs of network. To be more specific, our work will focus on processing problems in autonomous driving such as building a decoder network that can detect 3-D object detection with only one input and classify several objects. We will try to ameliorate lane lines performance as well as context of structures in drivable area segmentation.

## 5. ACKNOWLEDGEMENT

We would like to thank our colleagues in the Information Technology Specialization Department of FPT University, Hanoi, Vietnam for their critical and relevant comments on the manuscript; Colleagues in the English Department who have helped to polish the English text.

**CONFLICT OF INTEREST:** The process of writing and the content of the article does not give grounds for raising the issue of a conflict of interest.

**COMPLIANCE WITH ETHICAL STANDARDS:** This article is a completely original work of its authors; it has not been published before and will not be sent to other publications until the PRIA editorial board decides not to accept it for publication.

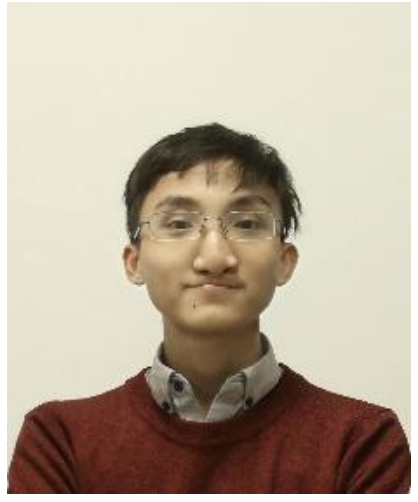

Vu Thanh Dat received his B.S.E degree in Computer Science from FPT University, Hanoi, Vietnam.

His current research interests include Artificial Intelligence, Computer Vision, Pattern Recognition, Machine Learning, and Autonomous Driving.

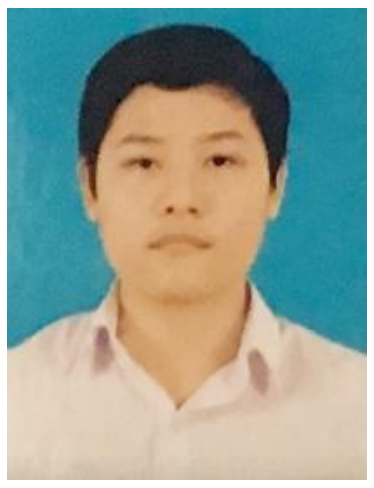

Ngo Viet Hoai Bao graduated from FPT University, Hanoi, Vietnam with a degree in Computer Science in 2022.

His current research interests include Artificial Intelligence and Deep Learning.

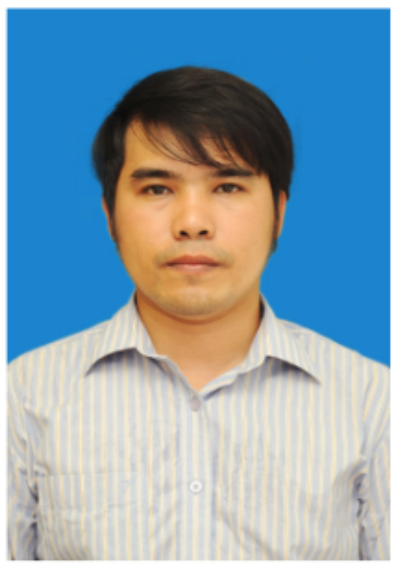

Phan Duy Hung received his Ph.D. degree from INP Grenoble France, in 2008. Since 2009, he has worked as a Lecturer, and served as the Head of Department and the Director of the Master Program in Software engineering at FPT University, Hanoi, Vietnam.

His current research interests include Digital Signal and Image processing, Internet of Things, BigData, Artificial Intelligence, Measurement and Control Systems and Industrial networking.